\documentclass[runningheads]{llncs}

\usepackage{eccv}

\usepackage{eccvabbrv}
\usepackage{wrapfig}
\usepackage{graphicx}
\usepackage{booktabs}
\usepackage[accsupp]{axessibility}  

\usepackage{hyperref}

\usepackage{orcidlink}

\usepackage{booktabs}
\usepackage{multirow}
\usepackage{rotating}
\usepackage{adjustbox}
\usepackage{bbm}
\def\eg{\emph{e.g}\onedot} 

\def\ie{\emph{i.e}\onedot} 

\usepackage{mathtools}
\usepackage{textcomp}
\usepackage{gensymb}
\usepackage{bm}
\usepackage[dvipsnames]{xcolor}
\usepackage{enumitem}

\newcommand{\x}[1]{{\color{red}{placeholder}}} 

\def\our{MeshVPR}

\newcommand{\myparagraph}[1]{\vspace{4pt}\noindent\textbf{#1}}

\begin{document}

\title{MeshVPR: Citywide Visual Place Recognition Using 3D Meshes} 

\titlerunning{MeshVPR}

\author{Gabriele Berton\inst{1} \and
Lorenz Junglas\inst{2} \and
Riccardo Zaccone\inst{1} \and
Thomas Pollok\inst{3} \and
Barbara Caputo\inst{1} \and
Carlo Masone\inst{1}}

\authorrunning{G.~Berton et al.}

\institute{Politecnico di Torino \and
Karlsruhe Institute of Technology \and
Fraunhofer IOSB}

\maketitle

\begin{abstract}
Mesh-based scene representation offers a promising direction for simplifying large-scale hierarchical visual localization pipelines, combining a visual place recognition step based on global features (retrieval) and a visual localization step based on local features. While existing work demonstrates the viability of meshes for visual localization, the impact of using synthetic databases rendered from them in visual place recognition remains largely unexplored. 
In this work we investigate using dense 3D textured meshes for large-scale Visual Place Recognition (VPR). We identify a significant performance drop when using synthetic mesh-based image databases compared to real-world images for retrieval. To address this, we propose MeshVPR, a novel VPR pipeline that utilizes a lightweight features alignment framework to bridge the gap between real-world and synthetic domains. MeshVPR leverages pre-trained VPR models and is efficient and scalable for city-wide deployments. We introduce novel datasets with freely available 3D meshes and manually collected queries from Berlin, Paris, and Melbourne. 
Extensive evaluations demonstrate that MeshVPR achieves competitive performance with standard VPR pipelines, paving the way for mesh-based localization systems. 
Data, code, and interactive visualizations are available at \small{\url{https://meshvpr.github.io/}}

\end{abstract}

\section{Introduction}
\label{sec:introduction}

Estimating the location of where a photo was taken 
based solely on its visual content is a staple of computer vision, and enables a number of applications 
ranging from augmented reality \cite{Fatouh_2021_VL_for_AR}, robotics localization \cite{Suomela_2024_Placenav} and assistive technology \cite{Cheng_2021_VL_impaired}.
It can be used as an alternative to GPS, where no signal nor internet connection is available or jammed \cite{Zeng_2017_underground_VPR, Chai_2023_underground_VPR}. Additionally, it can help with automatic localization of non-geotagged image and video footage, which can be useful during or after tragic events like the terror attacks of Paris in 2015, to enable a time-critical investigation instead of months of manual labour.

The approaches developed to tackle the problem depend on factors like the size of the localization map (\eg a small building vs a large city) and the required accuracy of the estimate (\eg, a coarse position or a precise pose).
In the most challenging scenarios, \ie, when the objective is the precise pose on a large map, 
it is common practice to rely on hierarchical solutions \cite{irschara2009structure,sattler2012image,sarlin2018leveraging,taira2018inloc,taira2019right,torii2019large,Panek2022ECCV}, which comprise two steps:
(1) a \emph{Visual Place Recognition} (\textbf{VPR}) step with global features, where methods such as NetVLAD \cite{Arandjelovic_2018_netvlad} are used to obtain a coarse prediction through efficient image retrieval, and 
(2) a \emph{Visual Localization} (\textbf{VL}) step with local features \cite{Sarlin_2020_superglue}, where the initial pose estimate is refined, often by matching keypoints from the image to be localized to a 3D model of the map.

\begin{wrapfloat}{figure}{O}{0.5\textwidth}
\vspace{-16pt}
\raggedleft
\includegraphics[width=0.5\textwidth]{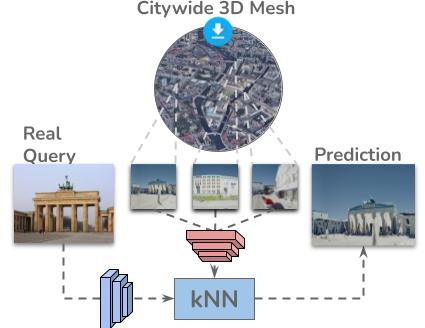}
\caption{\textbf{Mesh-based VPR.} Deploying a VPR model to a city
can be performed with a database of synthetically generated images from a 3D mesh.
While standard image retrieval techniques can be applied, a natural drop in results is due to the domain shift between real images (the query to be localized) and the synthetic database, requiring for the search of new solutions.
}
\vspace{-12pt}
\label{fig:teaser_motivation}
\end{wrapfloat}

Both VPR and VL have been thoroughly studied in literature, either independently or within hierarchical pipelines.
The vast majority of works relies on
creating a world map with RGB images, which are used 
in VPR as a retrieval database, and can then be employed to create a 3D point cloud of the map, with each point being associated to a feature vector.
Such point clouds are inherently tied to the model used to extract their features, limiting their flexibility and adaptability to new methods.

Using a different approach, a separate branch of literature demonstrates the viability of using dense 3D textured mesh models as scene representation for visual localization \cite{zhang2021reference, Sibbing_2013_sift_rendering, Qi_2014_vloc_on_aerial_mesh, Aubry_2016_vloc_w_meshes, Panek2022ECCV, panek2023visual}.
Overall, such methods find that the synthetic-to-real domain shift between the map and queries does not strongly impact the localization, noting that local features are robust to such visual changes.
Conversely,
this synthetic-to-real shift has been scarcely investigated on global features for VPR:
previous mesh-based localization works have either (i) used small scenes, skipping the VPR step altogether \cite{Qi_2014_vloc_on_aerial_mesh, Sibbing_2013_sift_rendering, Aubry_2016_vloc_w_meshes},
(ii) relied on a database of real images for retrieval, before a mesh-based post-processing \cite{Panek2022ECCV}, or
(iii) used retrieval on maps of limited dimension \cite{panek2023visual, vallone2022danish} and did not report a noticeable performance drop related to the domain shift.

Addressing this knowledge gap is crucial for enabling  the creation of fully mesh-based large-scale hierarchical localization pipelines -- an attractive direction due to the increasing availability of free 3D textured meshes for multiple cities \cite{Lei_2023_Assessing_3D_City_Models}.
This work is the first investigation on VPR using global features on citywide 3D meshes:
on a large-scale San Francisco dataset we find that a SOTA VPR model achieves a Recall@1 of 96.3\%  when using a real-images database, and drops to 76.9\% when using a database made of synthetic images (see \cref{fig:teaser_motivation}).

To bridge this gap, we propose a novel VPR model that excels in three key aspects: (i) it efficiently matches real-world photos to a database of synthetic images, (ii) it scales effectively for large-scale datasets, enabling city-wide deployments, and (iii) it delivers strong results, achieving performance competitive with standard real-world VPR pipelines.
We achieve this by developing a lightweight features alignment procedure that leverages pre-trained, state-of-the-art VPR models and adapts them to ensure consistent feature representations between real and synthetic domains. This paper details the complete pipeline, including training data preparation, features alignment, testing data acquisition from 3D meshes, and model deployment.
We then evaluate our pipeline over a number of novel datasets, which rely on freely available 3D meshes and manually collected sets of queries.
Finally, we demonstrate how our pipeline, called MeshVPR, achieves impressive results on each of the datasets, performing competitively with standard VPR pipelines using real-world database images.
To summarize, our contributions are:
\begin{itemize}[topsep=0pt,itemsep=-1ex,partopsep=1ex,parsep=1ex]
  \item {\our}, a novel mesh-based VPR pipeline that uses a lightweight, model-agnostic and hyper-parameters free features alignment step to compensate for the mismatch between real-world and generated images. {\our} leverages the broad availability of citywide 3D meshes as well as pretrained VPR models to enable quick deployment of a VPR system on a target city.
  \item The release of our three 
  test sets from Berlin, Paris, and Melbourne, with manually collected real-world queries. Additionally, we also release a set of synthetic images from San Francisco to be used together with the already existing SF-XL \cite{Berton_2022_cosPlace} dataset to perform features alignment, without the need to collect more real-world images.
  \item A thorough analysis of failure cases and open challenges, noting that despite the large domain gap MeshVPR is able to achieve excellent quantitative and qualitative results, leading the way for future research and for new mesh-based localization systems.
\end{itemize}

\section{Related work}
\label{sec:related_work}

\myparagraph{Hierarchical Visual Localization} pipelines \cite{sarlin2019coarse, Torii_2021_r_sf} rely on the combination of two steps: a visual place recognition (retrieval) step to obtain a coarse pose estimate, and a visual localization step, which aims at obtaining a precise camera pose.
While both tasks have been largely investigated for real-to-real localization, and recent works explored the possibility of mesh-based visual localization \cite{Panek2022ECCV, panek2023visual}, mesh-based visual place recognition is still largely unstudied, despite being crucial for scalable hierarchical pipelines.
Below we present a summary on (real-to-real) visual place recognition and syn-to-real localization works.

\myparagraph{Visual Place Recognition} 
Early VPR methods relied on representations obtained from hand-crafted local features, such as SIFT \cite{Lowe_2004_sift} and RootSIFT \cite{Arandjelovic_2012_rootSift}, while modern approaches have shifted to deep learning based approaches, where features are learned with CNNs \cite{Babenko_2014_neural_codes, Arandjelovic_2018_netvlad, Trivigno_2023_divideAndClassify, Schubert_2023_VPRtutorial} or visual transformer architectures \cite{Zhu_2023_r2former}.
Several works have proposed methods to condense these features into compact and discriminative global embeddings, \eg using pooling layers \cite{Razavian_2015_mac,Tolias_2016_rmac,Radenovic_2019_gem, Arandjelovic_2018_netvlad, Mereu_2022_seqvlad, Hausler_2019, Neubert_2021_HDC}, clustering based approaches~\cite{Zhang_2021_gated_netvlad,Kim_2017_crn,Peng_2021_appsvr}, MLPs \cite{Alibey_2023_mixvpr} or self-attention \cite{Zhu_2023_r2former}.
Most notably, to ensure that the model learns to extract specific features for urban VPR, Arandjelovic et al. \cite{Arandjelovic_2018_netvlad} proposed to train it on a dataset of StreetView images in a weakly supervised way.
Many following works built on top of NetVLAD, enhancing it with an attention module \cite{Kim_2017_crn}, a novel loss \cite{Liu_2019_sare}, or a self-supervised training strategy \cite{Ge_2020_sfrs}.
Recent state-of-the-art methods \cite{Alibey_2023_mixvpr, Berton_2023_EigenPlaces, izquierdo_2024_SALAD} rely on efficient training paradigms, and leverage large-scale training datasets such as GSV-cities \cite{Alibey_2022_gsvcities} and SF-XL \cite{Berton_2022_cosPlace}.

\myparagraph{Synthetic data for localization tasks}
The use of synthetic images has been explored for pose estimation with respect to a large indoor 3D map \cite{Taira_2018_inloc}. Aiming at addressing changes in appearances, \cite{Torii_2018_tokyo247} has explored view synthesis to align viewpoints of queries and database.
Despite the existence of numerous synthetic 3D meshes of cities \cite{deschaud2021pariscarla3d, Lei_2023_Assessing_3D_City_Models} and synthetically-generated environments \cite{Alberti_2020_idda, Lin_2023_InfiniCity, Li_2023_MatrixCity, Qui_2017_unrealcv, Pollok_2019_unrealgt, Shah_2018_AirSim,Dosovitskiy_2017_carla}, their usage for image localization has been little explored in recent years.
Among the few outliers, Panek et al \cite{Panek2022ECCV, panek2023visual} proposes the use of 3D meshes for visual localization, although still relying on real-world images for the retrieval step.
On a separate development, Vallone et al \cite{vallone2022danish} spearheaded the exploration of aerial images to render street-view images for place recognition, and used it to generate a database of 44k images for the coarse localization of real-world query images.
More in general, \cite{Moreau2021LENSLE} explored the use of synthetic data to improve the localization process, while others further train the model with real and synthetic data to improve feature representation \cite{Moreau2023CROSSFIRECR}.

This paper fills a noticeable gap in the literature, as it is the first work to (i) explore 3D meshes for citywide VPR, (ii) quantify the gap in results between using real images and synthetic images, (iii) use features alignment to efficiently adopt robust SOTA VPR models and (iv) provide a number of large-scale datasets comprising challenging manually-collected query images.

\begin{figure}[!t]
\centering
\includegraphics[width=0.99\columnwidth]{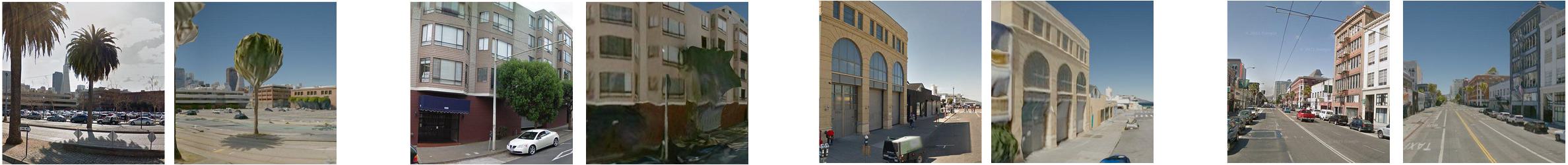}
\caption{\textbf{Pairs of real images and their synthetic counterpart.} Pairs like these are used for MeshVPR's features alignment.}
\label{fig:real_vs_syn}
\end{figure}

\section{Mesh-based Visual Place Recognition Pipeline}
\label{sec:pipeline}

\myparagraph{From standard to mesh-based VPR.}
VPR systems are usually implemented following an image retrieval approach. The images of the target city's database constitute the hypotheses for the incoming visual query, and predicting its location involves a similarity search in that database. In practice, the query's location is inferred from the most relevant matches of this retrieval process.
Formally, the database $(\mathcal{X}, \mathcal{Y})$ is composed of images $x \in \mathcal{X}$ annotated with a label $y \in \mathcal{Y}$ from the target city where the system needs to be deployed.
This pipeline consists of two steps~\cite{Masone_2021_survey}: first, the query image $x_q$ to be localized and all the images from the database $\mathcal{X}$ are mapped to a common embedding space determined by a model $f_\theta$ with learnable parameters $\theta$; then, a k-nearest neighbor search (kNN) is performed to retrieve  the database images most similar to the query $x_q$, and their labels.

\begin{wrapfloat}{figure}{O}{0.5\textwidth}
\centering
\vspace{-10mm}
    \includegraphics[width=0.45\columnwidth]{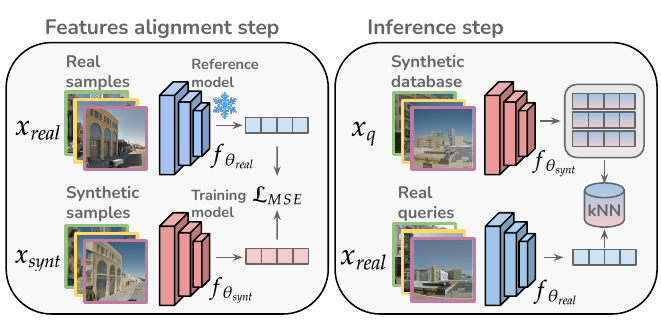}
    \caption{\textbf{At training time}
    the goal is to align features extracted by $f_{\theta_{real}}$ from real images to those extracted by $f_{\theta_{synt}}$ from synthetic images.
    \textbf{At testing time} features extracted from synthetic images are used to localize the query (a real image).
    }
    \label{fig:architecture}
\vspace{-6mm}
\end{wrapfloat}

\begin{figure*}[t!]
    \centering
    \includegraphics[width=0.99\textwidth]{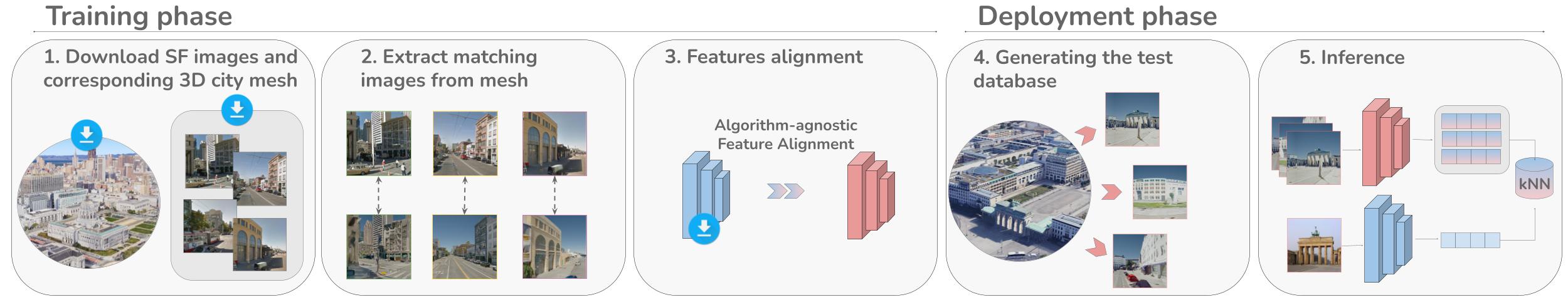}
    \caption{\textbf{Our proposed pipeline for mesh-based visual place recognition.} 
    The training phase consists in downloading training (real) images and the 3D mesh, generating their synthetic counterparts and specializing the synthetic model through feature alignment. 
    Once the training phase is completed, the deployment phase can take part on any target city: in this paper we show results on Berlin, Paris and Melbourne.
    }
    \vspace{-6mm}
    \label{fig:pipeline}
\end{figure*}

Motivated by the ever-growing availability of freely-available citywide 3D textured mesh models \cite{Lei_2023_Assessing_3D_City_Models},
and the recent advances in mesh-based visual localization \cite{Panek2022ECCV, panek2023visual}, our idea is to use such 3D meshes to obtain a synthetic database.
Note that these meshes are generally built from aerial images, which can not directly be used for VPR.
Formally, given a target city where we want to deploy a VPR system, we define 3D mesh as $\mathcal{M}_{target}$, and the database of images extracted from it as $(\mathcal{X}_{target}^{synt}, \mathcal{Y}_{target})$, the goal is to correctly match a given query image to a synthetic image from $\mathcal{X}_{target}^{synt}$ that depicts the same scene.
To allow VPR models to perform well in such conditions (\ie real query against a synthetic database), we devised a new \textit{mesh-based VPR pipeline} that can be applied to any existing VPR model $f_{\theta_{real}}$, trained on real-world data. Due to the appearance gap between real and synthetic images (see \cref{fig:real_vs_syn}), the features that $f_{\theta_{real}}$ extracts from a couple of real/synthetic images, even from the same viewpoint, are not well aligned.
To address this problem, we use a new model $f_{\theta_{synt}}$, that is initialized by setting $\theta_{synt} = \theta_{real}$ and then fine-tuned so that the embeddings $f_{\theta_{synt}}(x_{synt})$ and $f_{\theta_{real}}(x_{real})$ for a pair of matching synthetic/real images (\ie, from the same viewpoint) are well aligned.
This is akin to a teacher-student paradigm, where $f_{\theta_{real}}$ is the teacher and $f_{\theta_{synt}}$, although this differs with many teacher-student models in that (i) the two models have the same architecture and (ii) they are trained on images from different domains that share the exact same viewpoint.
Finally, the deployment is completed by using the specialized model $f_{\theta_{synt}}$ to extract the embeddings from the synthetic database $\mathcal{X}_{target}^{synt}$, whereas the model $f_{\theta_{real}}$ extracts the embeddings from the real-world queries. This idea is illustrated in \cref{fig:architecture} (right).
The overall deployment pipeline is a \emph{5-steps recipe}, as illustrated in \cref{fig:pipeline}. Remarkably, the features alignment (steps 1-3) needs to be performed only once to align the model $f_{\theta_{synt}}$, while any actual deployment only requires the database extraction from the 3D mesh of the target city.

\subsection{Step 1: Download images and mesh for the alignment}
\label{sec:step1}
The crucial part in our pipeline is the features alignment (described in \cref{sec:step3}), such that real and synthetic images are aligned in the feature space. To prepare for this, it is necessary to download a 3D mesh for any city for which a training set of real images is available. Notably, there is no need for this dataset and 3D mesh to
be from the target city(ies), and in most of our experiments we pick a different city from the target one.
Therefore, in this step, we simply download a dataset of real-world images $\mathcal{X}_{align}^{real}$ and a 3D mesh $\mathcal{M}_{align}$ from the same city.
The images $\mathcal{X}_{align}^{real}$ must be labeled both with GPS coordinates and heading, since orientation is crucial in the next step to generate synthetic images from the same viewpoint as the real ones.


\subsection{Step 2: Generate alignment images from mesh}
\label{sec:step2}
With downloaded the citywide 3D mesh and the corresponding training dataset, it is then possible to create a set of synthetic images $\mathcal{X}_{align}^{synt}$ from the mesh $\mathcal{M}_{align}$, such that they precisely mirror the images from $\mathcal{X}_{align}^{real}$ (see \cref{fig:real_vs_syn}).

This requires to have the latitude, longitude and heading (yaw) of each real image in $\mathcal{X}_{align}^{real}$, as they are needed to create views of the mesh from the corresponding viewpoints. Altitude, pitch and roll are also required to  accurately replicate the images, but these can be inferred from the physics information of the 3D mesh.
To accomplish that, we cast a ray from the real image location towards the ground and see where it intersects with the model.
This intersection point serves as a reliable estimate of the ground level. The views are then generated at a height of 2.5m, \ie, roughly the height of a typical car-mounted camera  used to collect VPR datasets \cite{Milford_2008_st_lucia, Maddern_2017_robotCar}. Additionally, the normal vector at the intersection is used to estimate pitch and roll.

Synthetic counterpart for real images can be generated from the mesh using any rendering engine: in our case, we rely on Unreal Engine and Cesium.


\subsection{Step 3: Features alignment}
\label{sec:step3}
In order to extract coherent embeddings from both real-world and synthetic images, we propose a strategy based on two expert models, $f_{\theta_{real}}$ and $f_{\theta_{synt}}$, for real and synthetic images respectively. 
The goal of this step is to ensure that $f_{\theta_{synt}}$ and $f_{\theta_{real}}$ produce similar embeddings from images that are taken from the same viewpoint.

To this end, we initialize $f_{\theta_{real}} = f_{\theta_{synt}}$ to any open-source pretrained VPR model, like CosPlace \cite{Berton_2022_cosPlace}, MixVPR \cite{Alibey_2023_mixvpr} or SALAD \cite{izquierdo_2024_SALAD}.
Secondly, we fine-tune $f_{\theta_{synt}}$ on the images $\mathcal{X}_{align}^{synt}$ to mirror the features extracted by $f_{\theta_{real}}$ from the corresponding images in $\mathcal{X}_{align}^{real}$. 
Formally, we optimize the following MSE loss:
\begin{equation}
\mathcal{L}_{MSE}= \frac{1}{N}\sum_{i=1}^{N}\left( f_{\theta_{real}}(x_{real,i}) -f_{\theta_{synt}}(x_{synt,i}) \right)^2
\end{equation}
where $N=|\mathcal{X}_{align}^{real}|=|\mathcal{X}_{align}^{synt}|$ and $x_{i}^{real}\in \mathcal{X}_{align}^{real}$, $x_{i}^{synt}\in \mathcal{X}_{align}^{synt}$ are pairs of real and synthetic images from the same viewpoint, and the parameters $\theta_{real}$ are frozen.

\underline{Note that}: (i) the training phase (steps 1-3) needs to be performed only once to create the model $f(\theta_{synt})$, which can then be deployed to any new city, and (ii) although these steps do require a real-world dataset $\mathcal{X}_{align}^{real}$ annotated with position and heading, it is possible to use existing datasets such as SF-XL~\cite{Berton_2022_cosPlace} without having to rely on images from the target city.


\subsection{Step 4: Generate the test database}
\label{sec:step4}
The next step consists in generating the database $(\mathcal{X}_{target}^{synt}, \mathcal{Y}_{target})$ for the target cities, \ie, the ones where we want to deploy the VPR system.
Given the 3D mesh $\mathcal{M}_{target}$ from the target city, we want to ensure that the database is constructed by generating relevant views, which could ideally match the camera viewpoints of the queries.

We generate street-view-like images from the 3D mesh by simulating the path that a camera-equipped car would take to collect a database.
This process begins with fetching OpenStreetMap (OSM) street data for the area covered by the 3D mesh $\mathcal{M}_{target}$. Then, a graph is constructed from the OSM data. For this graph we compute a path that visits every edge, this is accomplished by solving the Route Inspection Problem for the graph.
This continuous path is needed in order to ensure equal spacing between the individual sampling locations and reduce the sampling time.
The actual capture is performed by moving the camera along the path by a configurable distance (10 meters in our experiments, unless otherwise stated).
We note that sampling can easily be adapted to the specific application, \eg a database for drone localization can be sampled from simulated aerial viewpoints. This underscores the flexibility of our approach for scenarios that are different from street-level visual place recognition.


\subsection{Step 5: Inference}
\label{sec:step5}
Finally, we deploy the system using the generated test database $(\mathcal{X}_{target}^{synt}, \mathcal{Y}_{target})$, and the models $f_{\theta_{real}}$ and $f_{\theta_{synt}}$.
First, we extract the embeddings from each image in $\mathcal{X}_{target}^{synt}$ using $f_{\theta_{synt}}$.
Then, the inference simply consists in taking any unseen real-world query $x_q$, extracting its embedding $f_{\theta_{real}}(x_q)$, and compare it to the embeddings of the database images via kNN.
Once the nearest neighbor(s) is found, we can directly infer the query's position from its metadata, following standard visual place recognition \cite{Arandjelovic_2018_netvlad}.

\section{Test sets}
\label{sec:test_sets}

To empirically validate the soundness of MeshVPR, we built three test sets, each one consisting of a synthetic geo-tagged database, built following the method in \cref{sec:step4}, and a number of real-world queries.
The three test sets cover the cities of Berlin, Paris and Melbourne,
for which 3D meshes are freely available: the datasets characteristics are summarized in \cref{tab:3d_models}.
For Berlin and Paris, a part of the queries have been manually collected, and another part has been downloaded from Flickr and Wikimedia, which are commonly used sources for computer vision datasets \cite{Philbin_2007_oxford5k, Philbin_2008_paris6k, Berton_2022_cosPlace}.
Some examples are shown in the qualitative results in \cref{sec:experiments} and in the Supplementary.
Given the inaccuracies in GPS positioning (both in our manually collected queries, and Flickr and Wikimedia photos), we manually selected only a few hundred queries for the two European capitals, carefully removing any image for which the GPS label did not match its actual position.
For Melbourne we use as queries a subset of the Mapillary Street-Level Sequences dataset \cite{Warburg_2020_msls} collected in the city of Melbourne.

\begin{table*}[!t]
\caption{\textbf{Characteristics of 3D mesh models used in our experiments.} 
The 5 central columns refer to the 3D mesh. The 3 right-most columns refer to the images used, either generated from the mesh (for the database) or the queries.
San Francisco HQ and San Francisco LQ indicate the higher/lower quality ones, which we use to investigate how mesh resolution affects results.
DB stands for database.
*Only the subset of San Francisco XL (SF-XL) overlapping both 3D meshes is considered;
$^\dagger$Different parts of the mesh have different quality.
}
\centering
\begin{adjustbox}{width=0.999\linewidth}
\begin{tabular}{cc|ccccc|ccc}
\toprule
Dataset Type & City & Size (GB) & Size (sq km) & Resolution (cm/px) & Year & Provider & DB images (GB) & \# DB images & \# queries \\
\midrule

\multirow{3}{*}{Train sets} &
  San Francisco XL (real images) &-&-&-&-&-                                     & 55 & 9.2M & 134 \\
& San Francisco HQ & 36 & 16.4 &0.6; 2; 5~$^\dagger$& 2021     & Aerometrex      & 55 & 9.2M* & 134* \\
& San Francisco LQ & 8  & 16.4 & $\approx15$        & 2015-2022& Google          & 55 & 9.2M & 134 \\
\midrule
\multirow{3}{*}{Test sets} &
Berlin             & 71 & 22.7 & $\approx9$         & 2020     &Senate of Berlin & 41 & 1.3M & 255 \\
& Paris            & 11 & 25.3 & $\approx15$        &2015-2023 & Google          & 45 & 1.8M & 268 \\
& Melbourne        & 21 & 13.4 & 7.5                & 2018     &City of Melbourne& 11 & 394k &1249 \\

\bottomrule
\end{tabular}
\end{adjustbox}
\label{tab:3d_models}
\end{table*}

\section{Experiments}
\label{sec:experiments}
In this section we present quantitative and qualitative results on mesh-based VPR, starting with MeshVPR's implementation details (\cref{sec:implementation_details}),
analyzing results on mesh-based visual place recognition with standard VPR models with and without the integration of MeshVPR (\cref{sec:mesh-basedVPR}), analyzing the importance of mesh quality for the task (\cref{sec:3D_mesh_quality}), exploring the performance gap when using a real or a synthetic database (\cref{sec:syn2real_performance}), exploring strategies alternative to MeshVPR (\cref{sec:other_strategie}), and finally analyzing advantages and limitations of using real or synthetic data for image localization.

\subsection{Implementation details}
\label{sec:implementation_details}
\myparagraph{Features Alignment.}
Our feature alignment is performed 
by pairing two copies of the same VPR model (\eg, NetVLAD, CosPlace, MixVPR), both initialized with open-source pretrained weights.
One of the two models, which we refer to as $f_{\theta_{real}}$, has frozen weights $\theta_{real}$, while the weights $\theta_{synt}$ of $f_{\theta_{synt}}$ are fine-tuned for 50k iterations with a batch size of 32 and the Adam \cite{Kingma_2014_adam} optimizer with learning rate $1e-5$.
Training takes only 3 hours on just 4GB of VRAM when using the heaviest model considered (\ie SALAD) on a Nvidia RTX-4090 GPU.

As per the training set, we use the largest publicly available VPR dataset, namely San Francisco eXtra Large (SF-XL) \cite{Berton_2022_cosPlace} and its 3D mesh counterpart, with synthetic views extracted from the San Francisco HQ mesh (see \cref{tab:3d_models}) unless otherwise stated.

\myparagraph{Inference.}
At inference time, $f_{\theta_{synt}}$ is used to extract features from synthetic database, while the queries are processed through $f_{\theta_{real}}$. As in standard VPR \cite{Arandjelovic_2018_netvlad, Ge_2020_sfrs, Alibey_2022_gsvcities, Berton_2022_cosPlace, Alibey_2023_mixvpr, Berton_2023_EigenPlaces}, queries' features are matched against database ones by kNN algorithm (see \cref{fig:architecture}, right). 
We use a threshold of 100 meters for positives.


\begin{table*}[!t]
\caption{\textbf{Evaluating methods on mesh-based VPR.} For each method, we show their performance with and without MeshVPR on our three new datasets made of synthetic database and real queries.
}
\centering
\begin{adjustbox}{width=0.99\linewidth}
\begin{tabular}{l|ccccc c ccccc c ccccc}
\toprule
\multirow{2}{*}{Method} &
\multicolumn{5}{c}{Synt-Berlin} &&
\multicolumn{5}{c}{Synt-Paris} &&
\multicolumn{5}{c}{Synt-Melbourne} \\
\cline{2-6} \cline{8-12} \cline{14-18}
& R@1 & R@5 & R@10 & R@20 & R@100 &
& R@1 & R@5 & R@10 & R@20 & R@100 &
& R@1 & R@5 & R@10 & R@20 & R@100 \\
\midrule

Conv-AP                      & 37.3 & 49.4 & 56.1 & 60.8 & 75.7 & & 34.3 & 47.8 & 55.2 & 61.2 & 73.5 & & 18.7 & 30.7 & 39.3 & 49.7 & 75.6 \\
Conv-AP + MeshVPR (Ours)     & 41.6 & 57.6 & 62.7 & 71.4 & 84.3 & & 35.4 & 51.5 & 56.3 & 62.7 & 76.1 & & 25.0 & 43.0 & 51.9 & 60.7 & 82.5 \\
\midrule
CosPlace                     & 55.7 & 64.7 & 69.4 & 72.5 & 82.4 & & 49.6 & 59.0 & 61.9 & 65.3 & 76.1 & & 38.4 & 52.1 & 61.5 & 69.5 & 84.9 \\
CosPlace + MeshVPR (Ours)    & 63.9 & 74.1 & 80.0 & 85.9 & 92.2 & & 50.7 & 62.3 & 68.3 & 73.1 & 79.9 & & 49.5 & 63.4 & 72.1 & 78.1 & 91.2 \\
\midrule
MixVPR                       & 48.6 & 63.1 & 67.5 & 72.5 & 83.9 & & 45.1 & 56.7 & 59.3 & 66.4 & 76.5 & & 28.4 & 45.6 & 53.1 & 61.0 & 81.8 \\
MixVPR + MeshVPR (Ours)      & 60.4 & 73.3 & 79.2 & 82.4 & 91.0 & & 52.2 & 60.4 & 66.0 & 73.1 & 82.1 & & 38.9 & 53.5 & 60.8 & 68.4 & 89.5 \\
\midrule
EigenPlaces                  & 52.2 & 63.5 & 67.8 & 72.9 & 79.6 & & 44.8 & 56.0 & 61.9 & 66.0 & 73.5 & & 26.7 & 40.4 & 47.6 & 54.2 & 74.7 \\
EigenPlaces + MeshVPR (Ours) & 67.1 & 80.4 & 83.5 & 85.9 & 92.2 & & 48.5 & 59.3 & 64.6 & 69.8 & 81.0 & & 47.6 & 64.1 & 70.3 & 75.6 & 89.6 \\
\midrule
SALAD                        & 69.8 & 79.6 & 83.1 & 84.7 & 93.3 & & 53.7 & 68.3 & 71.3 & 74.6 & 83.6 & & 59.1 & 73.9 & 78.8 & 84.3 & 94.6 \\
SALAD + MeshVPR (Ours)       & \textbf{82.0} & \textbf{89.4} & \textbf{92.2} & \textbf{92.9} & \textbf{95.7} & & \textbf{63.8} & \textbf{76.5} & \textbf{80.2} & \textbf{81.3} & \textbf{84.7} & & \textbf{69.1} & \textbf{81.7} & \textbf{86.8} & \textbf{92.2} & \textbf{97.8} \\

\bottomrule
\end{tabular}
\end{adjustbox}
\label{tab:main_table}
\end{table*}

\subsection{Localizing Real Queries on a Synthetic Database}
\label{sec:mesh-basedVPR}
\begin{figure}[!t]
    \begin{center}
        
    \includegraphics[width=0.99\columnwidth]{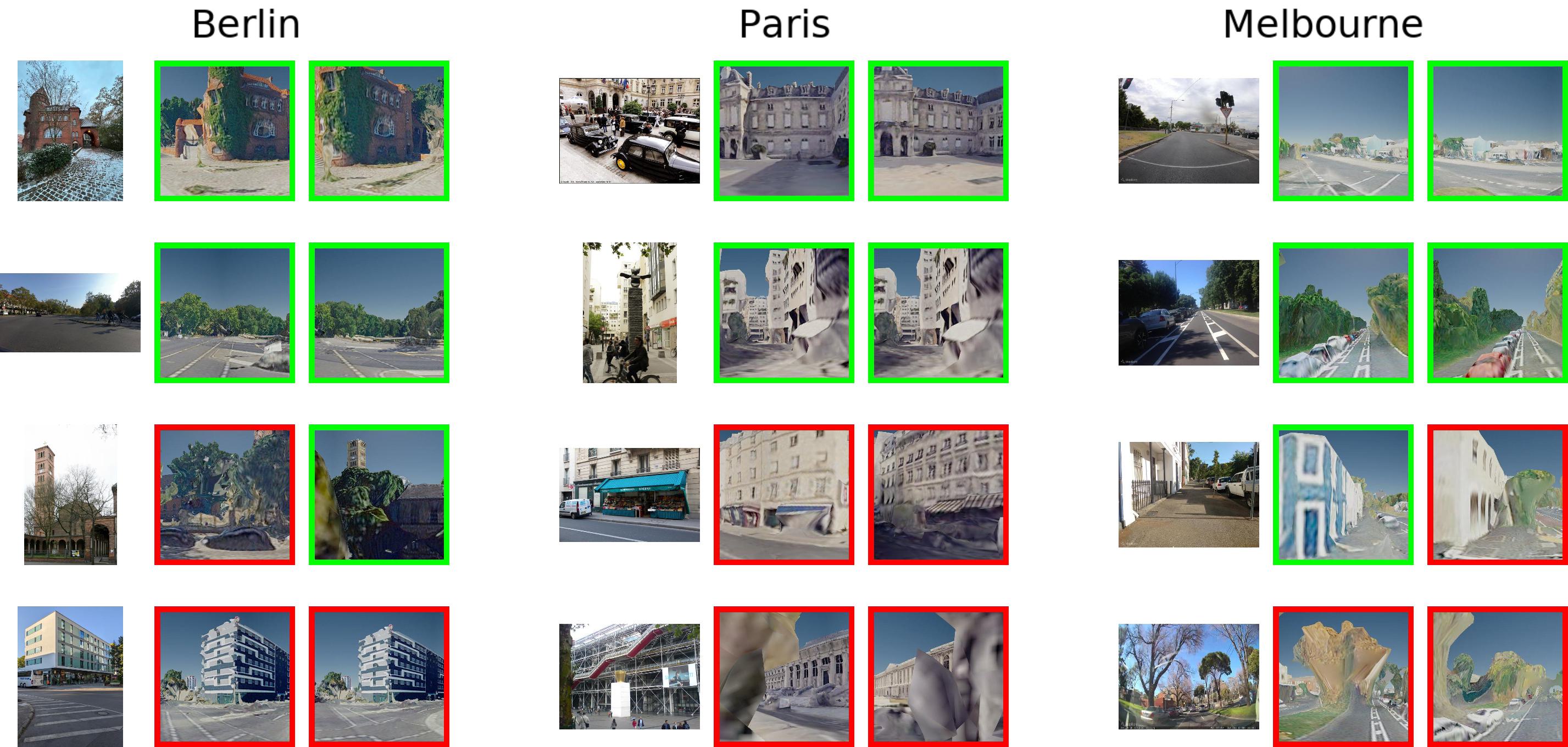}
    \end{center}
    \caption{\textbf{Predictions with best MeshVPR model}, namely SALAD + MeshVPR. Each triplet represents a query and its top 2 predictions, which are bounded in green if positive and red if negative. Qualitative examples help understand the results from \cref{tab:main_table}: Paris is challenging due to low quality meshes, and Melbourne is challenging due to wide open spaces.
    Interestingly, we note that the model learns to overcome long-term temporal changes (snow and winter/summer foliage in top-left query), occlusions (first two queries from Paris) and perspective changes (third query from Melbourne).
    A large number of (higher resolution) qualitative results are shown in the Supplementary.
    }
    \label{fig:preds_ber_paris_melb}
\end{figure}

Firstly, we investigate the feasibility of localizing real-world queries through a synthetic database.
To this end, we use a number of standard VPR methods, to see how results change when MeshVPR is applied.
Note that MeshVPR can be applied on top of any VPR model, which we showcase by testing SOTA single-stage VPR methods since 2022 \cite{Alibey_2022_gsvcities, Berton_2022_cosPlace, Alibey_2023_mixvpr, Berton_2023_EigenPlaces, izquierdo_2024_SALAD} with and without MeshVPR.
As it is possible to notice in \cref{tab:main_table}, even methods that have been proved to be robust to diverse training and test distributions, like MixVPR and EigenPlaces, perform poorly in the mesh-based VPR setting, confirming that the challenges posed by using synthetic images are beyond the generalization capabilities of any specific algorithm.
MeshVPR allows to achieve impressive results, greatly increasing the respective baselines.
Qualitative results in \cref{fig:preds_ber_paris_melb} give an insight into what MeshVPR learns during its feature alignment stage:
most notably, the model is able to overcome long-term temporal changes, seasonal changes, occlusions and perspective changes.
We note also how it is able to match features that have highly different appearance within the two domains, for example queries with trees produce predictions with trees, despite real and synthetic trees having little visual similarity.

\subsection{How 3D mesh quality affects results}
\label{sec:3D_mesh_quality}
\begin{figure}[!t]
\centering
    \includegraphics[width=0.99\linewidth]{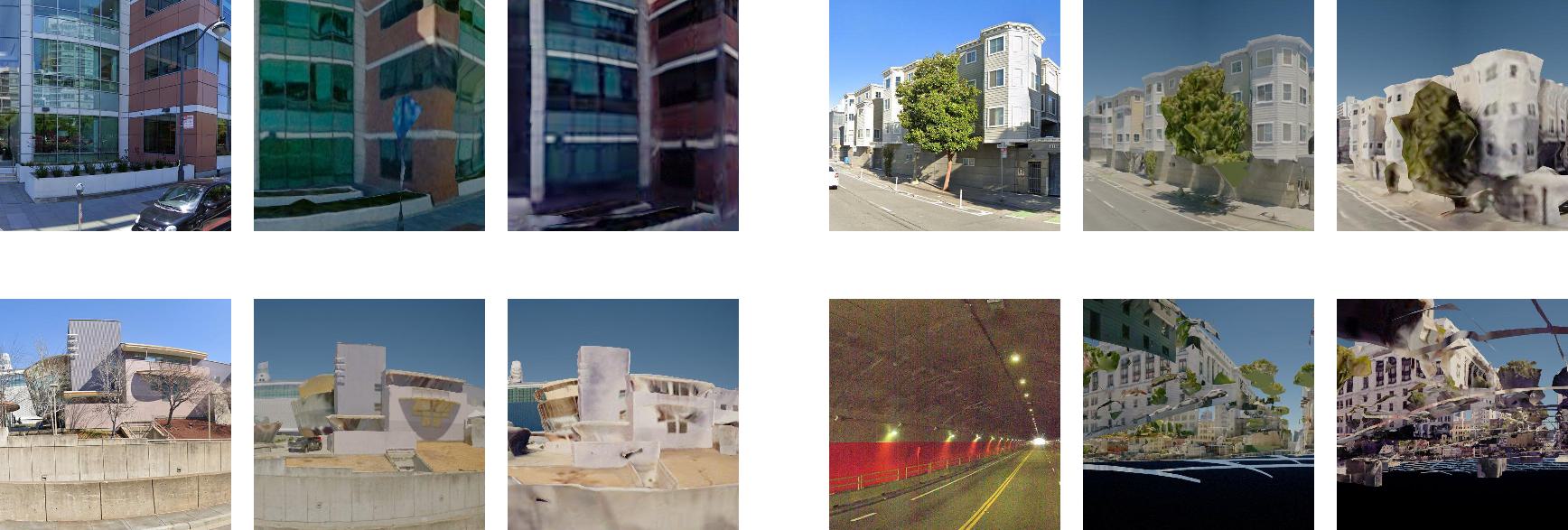}
    \caption{
    \textbf{Triplets of real, synthetic from HQ mesh, and synthetic from LQ mesh.} These triplets allow to qualitatively understand how the quality of the mesh influences the generated images and results. The bottom-right triplet provides a examples of synthetic images with artifacts. They occur when the real image was taken in a covered area i.e tunnel or tree cover, and the viewpoint is \emph{within} the mesh. Examples with such artifacts account for less than 1\% of the dataset.
    }
    \label{fig:real_hq_lq}
\end{figure}

Given that citywide 3D meshes play a central role in MeshVPR, we investigate the effect of their quality on the mesh-based VPR system.
For this purpose, we take the two overlapping 3D mesh of San Francisco LQ and San Francisco HQ (cf. \cref{tab:3d_models}) and use them both to generate synthetic views of the city (cf. \cref{fig:real_hq_lq}).

We provide results when the features alignment is performed on each of these datasets (whereas the real images for features alignment are unchanged).
To ensure that the areas of features alignment and testing do not overlap, we split the datasets into two non-overlapping halves (respectively north and south of latitude 37.78°) that are used as training and test sets, respectively. Note that we do not need validation as MeshVPR does not have hyperparameters, nor we perform early stopping.
The choice of using geographically disjoint train/test sets is in accordance to the typical VPR approach \cite{Arandjelovic_2018_netvlad, Berton_2021_svox, Warburg_2020_msls}, where non-overlapping train/test sets are used.
As queries, we use those from SF-XL \cite{Berton_2022_cosPlace} that are within the test area.

\begin{figure}[!t]
\begin{minipage}{0.44\textwidth}
    
\begin{adjustbox}{width=0.99\linewidth}
\begin{tabular}{lcc|ccccc}
\toprule
Method & Trained on & Tested on & R@1 & R@10 & R@100 \\
\midrule
\multirow{4}{*}{MixVPR}      & \multirow{2}{*}{LQ DB} & LQ DB & 59.7 & 74.6 & 91.0 \\
                             &                        & HQ DB & 59.0 & 77.6 & 93.3 \\
\cmidrule(l){2-3}            & \multirow{2}{*}{HQ DB} & LQ DB & 53.0 & 69.4 & 91.8 \\
                             &                        & HQ DB & 70.9 & 83.6 & 94.0 \\
\cmidrule(l){1-3}
\multirow{4}{*}{EigenPlaces} & \multirow{2}{*}{LQ DB} & LQ DB & 75.4 & 82.1 & 94.0 \\
                             &                        & HQ DB & 80.6 & 88.1 & 94.0 \\
\cmidrule(l){2-3}            & \multirow{2}{*}{HQ DB} & LQ DB & 67.9 & 81.3 & 94.8 \\
                             &                        & HQ DB & 85.8 & 91.8 & 97.0 \\
\cmidrule(l){1-3}
\multirow{4}{*}{SALAD}       & \multirow{2}{*}{LQ DB} & LQ DB & 76.1 & 90.3 & 97.8 \\
                             &                        & HQ DB & 87.3 & 91.0 & 97.8 \\
\cmidrule(l){2-3}            & \multirow{2}{*}{HQ DB} & LQ DB & 73.1 & 86.6 & 97.8 \\
                             &                        & HQ DB & 88.8 & 94.8 & 97.8 \\
                             
\bottomrule
\end{tabular}
\end{adjustbox}
\label{tab:database_quality}

\end{minipage}
\qquad
\begin{minipage}{0.55\textwidth}
    \input{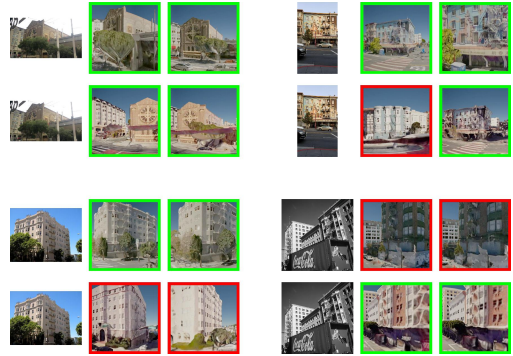}
\end{minipage}
\caption{\textbf{Results with MeshVPR on High Quality (HQ) and Low Quality (LQ) meshes.} Quantitative results (left) indicate a strong correlation between results and mesh quality. All results on the table are computed with MeshVPR applied to different VPR models. 
Qualitative results (right) visually show how predictions are affected by the synthetically generated images.
For each one of the 4 queries (\ie the images without green/red boxes) we show the top-2 candidates with SALAD+MeshVPR on the high-quality (HQ) database (top row for each query) and the top-2 candidates with the low-quality (LQ) database.
}
\label{fig:database_quality}
\end{figure}

Results are shown in \cref{fig:database_quality}, and they prove that the mesh quality has a huge impact on the results. Using the HQ (high quality) mesh (for both train and test) leads to a 15.7 points R@1 improvement over the LQ (low quality) mesh.
Results show that the scores obtained when testing on the LQ database benefit if the training is also performed on LQ data. For this reason, we will release both the models trained with HQ and LQ data.

Albeit somewhat predictable, this result is promising for the development of mesh-based VPR: in fact, as the quality (and availability) of these city-wide 3D meshes is steadily increasing, so will the results of MeshVPR.

\subsection{Bridging the Syn2Real performance gap}
\label{sec:syn2real_performance}
\begin{figure}
\begin{minipage}{0.48\textwidth}

\begin{adjustbox}{width=0.99\linewidth}
\begin{tabular}{ll|ccc}
\toprule
Scenario & Method & R@1 & R@10 & R@100 \\
\midrule
\multirow{5}{*}{Standard VPR}     & Conv-AP                      & 73.1 & 91.0 & 96.3 \\
                                  & CosPlace                     & 88.1 & 93.3 & 98.5 \\
                                  & MixVPR                       & 88.1 & 94.8 & 97.0 \\
                                  & EigenPlaces                  & 92.5 & 96.3 & 98.5 \\
                                  & SALAD                        & 96.3 & 98.5 & 99.3 \\
\midrule
\multirow{10}{*}{Mesh-based VPR}  & Conv-AP                      & 31.3 & 51.5 & 78.4 \\
                                  & CosPlace                     & 67.2 & 79.9 & 88.8 \\
                                  & MixVPR                       & 48.5 & 70.1 & 90.3 \\
                                  & EigenPlaces                  & 64.9 & 79.9 & 89.6 \\
                                  & SALAD                        & 76.9 & 88.8 & 97.8 \\
\cmidrule(l){2-5}
                                  & Conv-AP + MeshVPR (Ours)     & 51.5 & 74.6 & 93.3 \\
                                  & CosPlace + MeshVPR (Ours)    & 80.6 & 86.6 & 93.3 \\
                                  & MixVPR + MeshVPR (Ours)      & 70.9 & 83.6 & 94.0 \\
                                  & EigenPlaces + MeshVPR (Ours) & 85.8 & 91.8 & 97.0 \\
                                  & SALAD + MeshVPR (Ours)       & 88.8 & 94.8 & 97.8 \\

\bottomrule
\end{tabular}
\end{adjustbox}

\end{minipage}
\qquad
\begin{minipage}{0.48\textwidth}
    \input{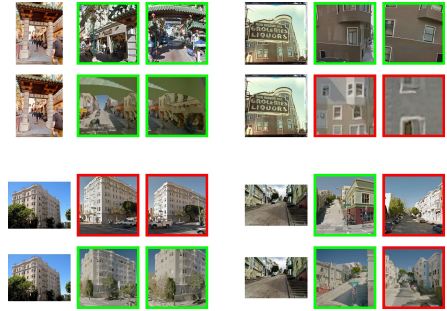}
\end{minipage}
\caption{\textbf{Quantitative and qualitative results with a real vs a synthetic database.} Quantitative results (left) show are performed with SOTA VPR models on a real-world database (top-5 rows), and mesh-based database with and without MeshVPR. Qualitative results (right) show examples of 4 queries with their top-2 predictions, with the prediction bounded in green/red if correct/wrong. For each query we show predictions with SALAD on the real DB (top), and SALAD+MeshVPR on the synthetic DB (bottom).}
\label{tab:synt_vs_real_db}
\end{figure}

Extensive validation presented in the above section demonstrates that MeshVPR makes each VPR algorithm obtain satisfactory performances on the mesh-based VPR setting.
A natural question arises: \textit{to what extent do we recover the performance we would have obtained if a database of real images had been available?}

To answer this, we conducted an experiment on both real and synthetic versions of the San Francisco dataset, again splitting the datasets along latitude 37.78° in training and test sets, to ensure that MeshVPR's features alignment is not performed on the database samples.
We compare the performance of state-of-the-art VPR models both when using a database of real-images, and when using the synthetic database.
Results in \cref{tab:synt_vs_real_db} show that, while a gap does still exist, MeshVPR recovers much of the performance loss caused by the use of the synthetic database. Most notably, the best state-of-art VPR model (SALAD) experiences a drop in Recall@5 (R@5) $<4\%$: the performance of SALAD with a real-world database is 97.8\%, opposed to 94.0\% of SALAD+MeshVPR on a synthetic database.



\subsection{Comparing MeshVPR with other strategies}
\label{sec:other_strategie}
Results from \cref{tab:other_baselines} demonstrate the benefit of MeshVPR's  two-models strategy, with one model extracting features from real images and the other extracting aligned features from synthetic images.
On the other hand, other solutions only alleviate the problem and do not achieve competitive results with MeshVPR.

\begin{table}[!t]
\caption{\textbf{Using metric learning losses to train a model on a dataset containing both synthetic and real images.} All methods start from a pretrained ResNet50-based EigenPlaces model. Other methods apply a "standard" fine-tuning of the model (in this case using both synthetic and real images), whereas MeshVPR uses the features alignment.
}
\centering
\begin{adjustbox}{width=0.99\textwidth}
\begin{tabular}{c|ccccc c ccccc c ccccc}
\toprule
\multirow{2}{*}{Method} &
\multicolumn{5}{c}{Synt-Berlin} & &
\multicolumn{5}{c}{Synt-Paris} & &
\multicolumn{5}{c}{Synt-Melbourne} \\
\cline{2-6} \cline{8-12} \cline{14-18}
& R@1 & R@5 & R@10 & R@20 & R@100 &
& R@1 & R@5 & R@10 & R@20 & R@100 &
& R@1 & R@5 & R@10 & R@20 & R@100 \\
\midrule
Contrastive loss                                      & 35.9 & 46.2 & 52.1 & 55.7 & 68.6 && 32.8 & 44.0 & 48.3 & 52.8 & 66.2 && 19.8 & 29.4 & 34.8 & 41.7 & 61.5 \\
Triplet loss as in \cite{vallone2022danish}           & 34.1 & 44.0 & 48.8 & 54.7 & 68.6 && 32.0 & 43.3 & 47.5 & 53.6 & 64.9 && 20.3 & 31.0 & 37.2 & 43.8 & 64.0 \\
Lifted loss \cite{Song_2016_liftedLoss}               & 38.0 & 48.5 & 53.7 & 59.1 & 73.3 && 34.5 & 45.6 & 50.8 & 55.2 & 65.2 && 22.6 & 35.1 & 41.8 & 47.8 & 67.9 \\
NTXent loss \cite{vandenOord_2018_NTXentLoss_InfoNce} & 37.6 & 48.5 & 54.5 & 59.0 & 72.3 && 34.1 & 47.1 & 50.9 & 55.6 & 66.6 && 23.4 & 35.5 & 41.4 & 49.2 & 68.0 \\
FastAP loss \cite{Cakir_2019_fastAP}                  & 37.3 & 48.2 & 54.2 & 60.1 & 72.4 && 34.2 & 45.8 & 50.6 & 54.6 & 66.5 && 24.5 & 37.6 & 43.9 & 50.1 & 68.6 \\
MultiSimilarity \cite{Wang_2019_multi_similarity_loss}& 40.4 & 50.6 & 57.8 & 63.3 & 77.5 && 37.4 & 47.8 & 52.4 & 57.1 & 68.5 && 24.5 & 35.5 & 41.7 & 49.4 & 69.1 \\
Circle loss \cite{Sun_2020_CircleLoss}                & 33.2 & 42.1 & 48.2 & 53.2 & 65.9 && 32.9 & 43.2 & 47.8 & 53.0 & 63.8 && 18.4 & 28.1 & 32.8 & 39.0 & 59.1 \\
SupCon loss \cite{Khosla_2020_SupCon}                 & 35.0 & 45.7 & 51.5 & 56.3 & 68.4 && 33.0 & 45.1 & 50.3 & 53.8 & 64.6 && 20.9 & 31.0 & 37.3 & 43.2 & 62.3 \\
\midrule
\textbf{MeshVPR (Ours) }  & \textbf{67.1} & \textbf{80.4} & \textbf{83.5} & \textbf{85.9} & \textbf{92.2} & & \textbf{48.5} & \textbf{59.3} & \textbf{64.6} & \textbf{69.8} & \textbf{81.0} & & \textbf{47.6} & \textbf{64.1} & \textbf{70.3} & \textbf{75.6} & \textbf{89.6} \\
\bottomrule
\end{tabular}
\end{adjustbox}
\label{tab:other_baselines}
\end{table}



\subsection{Training on fewer images}
\label{sec:training_on_fewer_images}
The San Francisco datasets (XL, HQ, and LQ) that we use for MeshVPR's features alignment all have exactly the same number of images from the same viewpoints, and comprise a large number of over 9.2M images each.
Whereas so many images are necessary to provide a proper measure of how changing the database's domain (\ie real or synthetic) for inference affects the results (see \cref{tab:synt_vs_real_db}), in this section we aim at understanding if so many images are actually necessary for the features alignment.
We report results in \cref{tab:training_on_fewer_images}, where we can see that using a much smaller training set of 100 k images leads to competitive results to using the full dataset.
Also, as detailed in \cref{sec:implementation_details}, note that in practice we train for 50k iterations with a batch size of 32, \ie the model sees 1.6M images and never actually goes through the entire dataset (we found convergence to be fully reached by 50k iterations).

\begin{table*}[!t]
\caption{\textbf{Training on fewer images.} Using the entire dataset (9.2 M pairs of images) for features alignment does not show large improvement, while using just 100 k images achieves almost the same results.
}
\centering
\begin{adjustbox}{width=0.99\linewidth}
\begin{tabular}{c|ccccc c ccccc c ccccc}
\toprule
\# training &
\multicolumn{5}{c}{Synt-Berlin} &&
\multicolumn{5}{c}{Synt-Paris} &&
\multicolumn{5}{c}{Synt-Melbourne} \\
\cline{2-6} \cline{8-12} \cline{14-18}
images & R@1 & R@5 & R@10 & R@20 & R@100 &
& R@1 & R@5 & R@10 & R@20 & R@100 &
& R@1 & R@5 & R@10 & R@20 & R@100 \\
\midrule

10 k  & 77.6 & 86.3 & 89.4 & 91.8 & 96.1 & & 59.3 & 72.8 & 75.7 & 81.0 & 84.3 & & 67.3 & 82.8 & 87.0 & 92.6 & 98.2 \\
100 k & 81.2 & 88.6 & 90.2 & 92.2 & 95.7 & & 61.2 & 76.1 & 78.4 & 79.5 & 84.7 & & 70.5 & 82.1 & 87.3 & 92.3 & 97.9 \\
1 M   & 80.0 & 87.5 & 90.2 & 93.3 & 96.1 & & 60.1 & 73.5 & 77.6 & 81.0 & 85.1 & & 69.7 & 80.6 & 85.6 & 90.9 & 97.8 \\
9.2 M & 82.0 & 89.4 & 92.2 & 92.9 & 95.7 & & 63.8 & 76.5 & 80.2 & 81.3 & 84.7 & & 69.1 & 81.7 & 86.8 & 92.2 & 97.8 \\

\bottomrule
\end{tabular}
\end{adjustbox}
\label{tab:training_on_fewer_images}
\end{table*}


\subsection{Limitations and advantages of mesh-based VPR}
Standard VPR (with real images for a database) and mesh-based VPR come each with their own strengths and weaknesses.
The main strength of standard VPR is the performance that SOTA models achieve even on large-scale datasets, as shown by the results in \cref{tab:synt_vs_real_db}, (and \cite{Alibey_2023_mixvpr, Berton_2023_EigenPlaces, izquierdo_2024_SALAD}).
Mesh-based VPR, having to overcome the strong domain shifts between database and query images, results in lower performance  even with the strong improvements given by MeshVPR.
On the other hand, mesh-based VPR provides a number of advantages:
\begin{enumerate}[topsep=0pt,itemsep=-1ex,partopsep=1ex,parsep=1ex]
    \item Higher potential for collecting data in challenging scenarios, like dangerous or remote locations, inaccessible to cars because of factors like damaged infrastructure, but accessible to flying drones.
    \item 3D citywide meshes provide a dense model of the city, covering virtually any outdoor location, whereas standard street-view based datasets cover only areas adjacent to roads.
    \item Fewer privacy issues, due to features like faces and car plates being absent in the mesh. For real-world images, these can cause issues, and blurring is often required when using such images.
    \item Increased flexibility in image sampling. 3D meshes allow simulating environmental factors, such as changes of appearance, light and weather conditions, which are difficult or expensive to capture in the real-world.
\end{enumerate}

\section{Conclusions and future works}
\label{sec:conclusions}

In this work, we propose the task of citywide mesh-based visual place recognition, which aims at obtaining scalable solutions to the problem of image localization, relying solely on a large 3D mesh.
We create three new datasets, to quantify the performance drop of existing models on this new task.
We then develop a simple yet effective technique, called MeshVPR, to align two models to the same feature space -- one model taking real images as inputs, and the other one taking synthetic images.
MeshVPR can be trained on any dataset of real and synthetic image pairs, on top of any existing VPR model, showing impressive results and strong flexibility.

We believe that the novel problem setup introduced in this paper opens countless future directions. For example:
(1) using a full mesh-based localization pipeline based on MeshVPR plus visual localization methods as in \cite{Panek2022ECCV, panek2023visual}, (2) generating synthetic images from multiple domains (e.g. synthetic night images), (3) simulate more viewpoints (e.g. from sidewalks or parks), (4) perform drone-based localization with a synthetic set of aerial images and (5) apply style transfer to transform all the images to a single domain.
In conclusion, citywide mesh-based VPR is an exciting new line of research which could unleash its full potential in allowing ubiquitous localization in practical scenarios.

\paragraph{\textbf{Acknowledgements.}}
\small{We acknowledge the Cineca award under the Iscra initiative, for the availability of high performance computing resources.
This work was supported by CINI.
Project supported by ESA Network of Resources Initiative.
This study was carried out within the project FAIR - Future Artificial Intelligence Research - and received funding from the European Union Next-GenerationEU (Piano nazionale di ripresa e resilienza (PNRR) – missione 4 componente 2, investimento 1.3 – D.D. 1555 11/10/2022, PE00000013 - CUP:
E13C22001800001). This manuscript reflects only the authors’ views and opinions, neither the European Union nor the European Commission can be considered responsible for them.
European Lighthouse on Secure and Safe AI – ELSA, Horizon EU Grant ID: 101070617
}

%
%
\bibliographystyle{splncs04}
\bibliography{main}

\end{document}


\title{MeshVPR: Citywide Visual Place Recognition Using 3D Meshes - Supplementary material} 

\titlerunning{MeshVPR: Citywide Visual Place Recognition Using 3D Meshes - Supplementary material}

\author{Gabriele Berton\inst{1} \and
Lorenz Junglas\inst{2} \and
Riccardo Zaccone\inst{1} \and
Thomas Pollok\inst{3} \and
Barbara Caputo\inst{1} \and
Carlo Masone\inst{1}}

\authorrunning{G.~Berton et al.}

\institute{Politecnico di Torino \and
Karlsruhe Institute of Technology \and
Fraunhofer IOSB}

\maketitle

\setcounter{page}{1}

\section{Examples of training and test images}

\begin{figure*}[t!]
\centering
\includegraphics[width=0.99\textwidth]{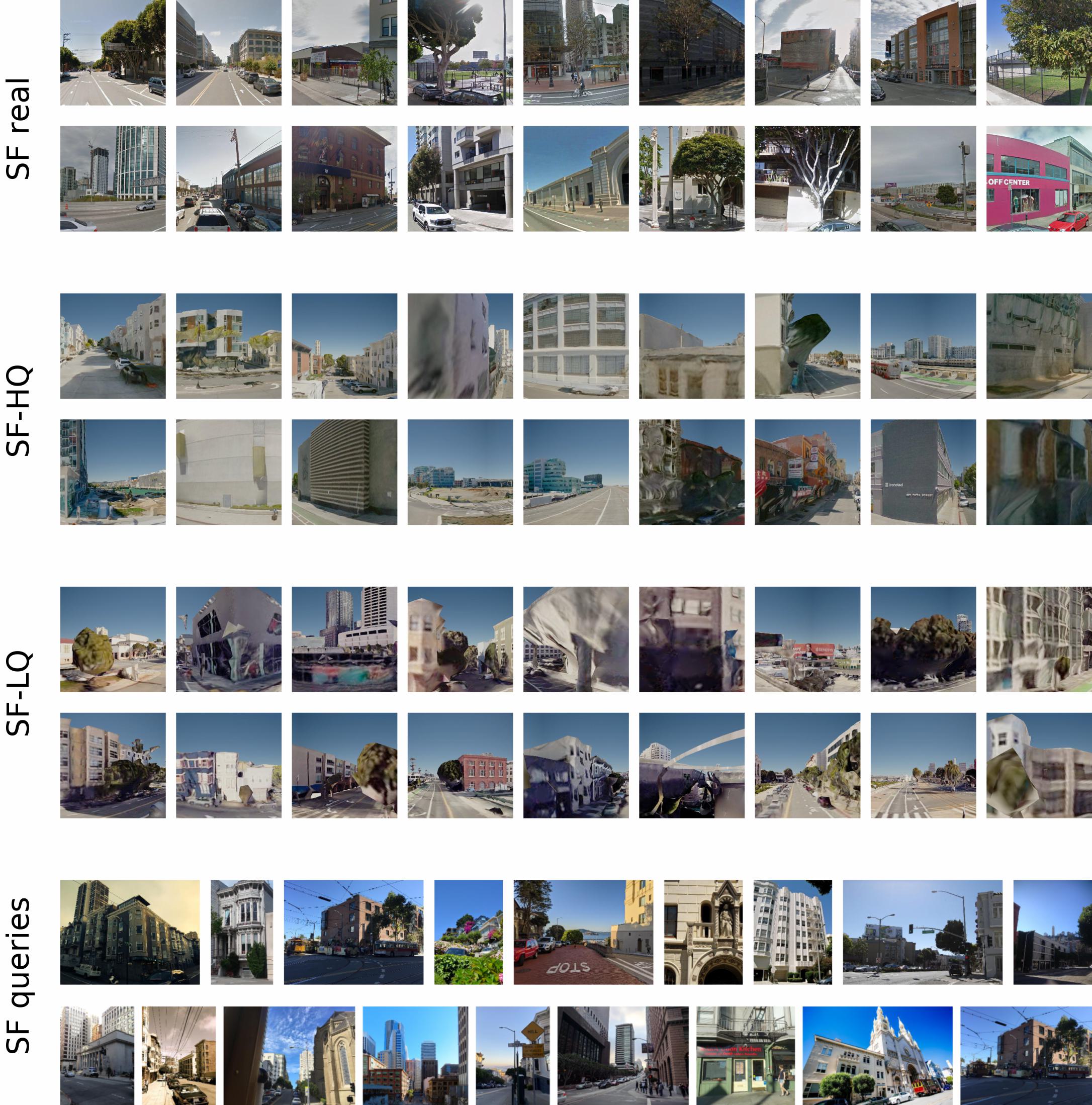}
\caption{\textbf{Examples of images from the datasets from San Francisco}, namely the real database, the High Quality (HQ) synthetic database, the Low Quality (LQ) synthetic database and the queries.}
\label{fig:supp_qualitatives_sf}
\end{figure*}

\begin{figure*}[t!]
\centering
\includegraphics[width=0.99\textwidth]{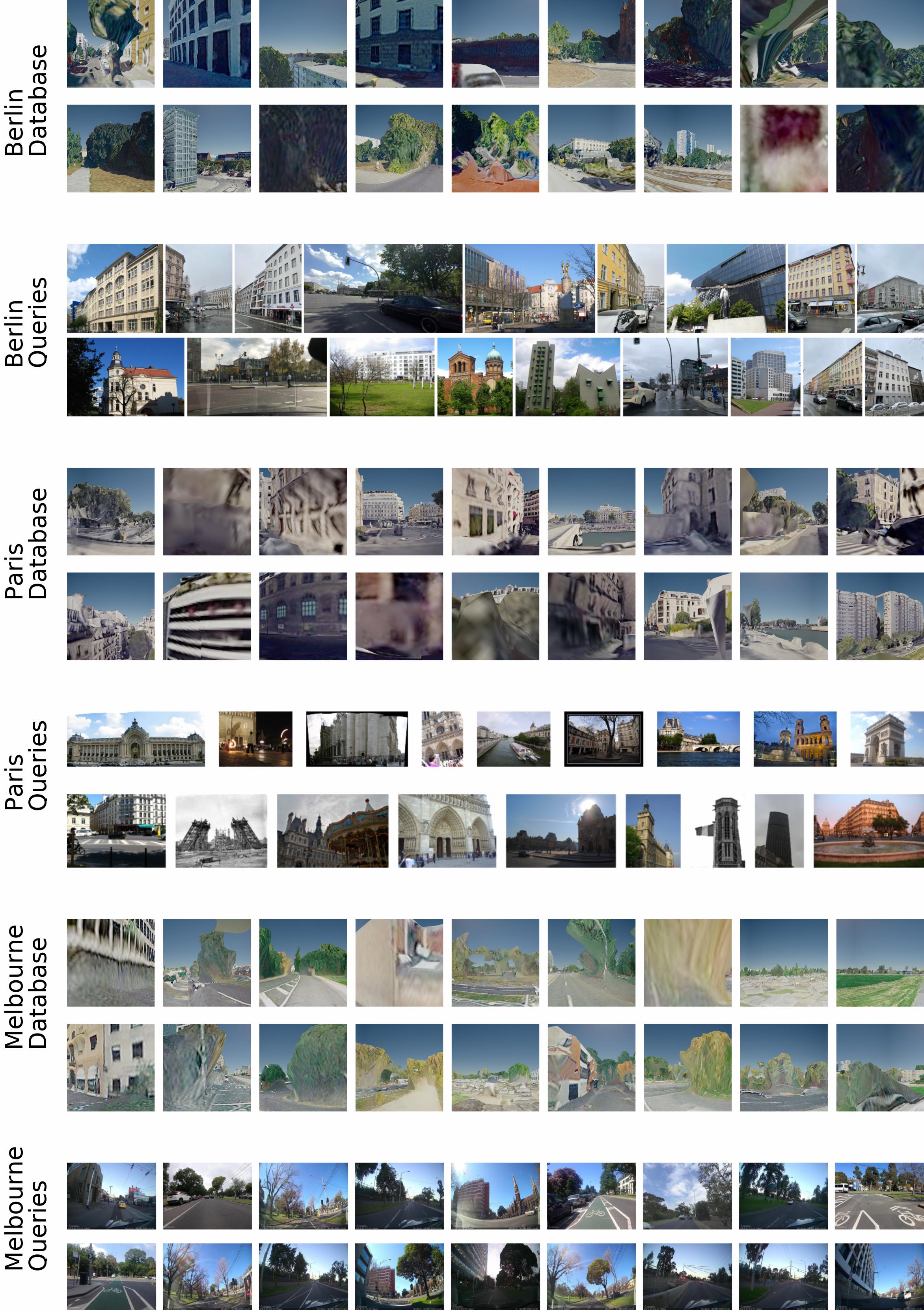}
\caption{\textbf{Examples of synthetic database and real queries from the datasets of Berlin, Paris and Melbourne.}}
\label{fig:supp_qualitatives_others}
\end{figure*}

In this Supplementary we present a large number of randomly chosen images from each of the training and test sets, shown in 
\cref{fig:supp_qualitatives_sf} (for the San Francisco datasets) and 
\cref{fig:supp_qualitatives_others} (for the datasets from Berlin, Paris and Melbourne).
These images allow to visually assess the large distribution gap between the various datasets, and, most importantly, between the synthetic and real images.

\section{Further Qualitative Results and Failure Cases}
\label{sec:supp_qualitative}
\begin{figure*}[t!]
\centering
\includegraphics[width=0.99\textwidth]{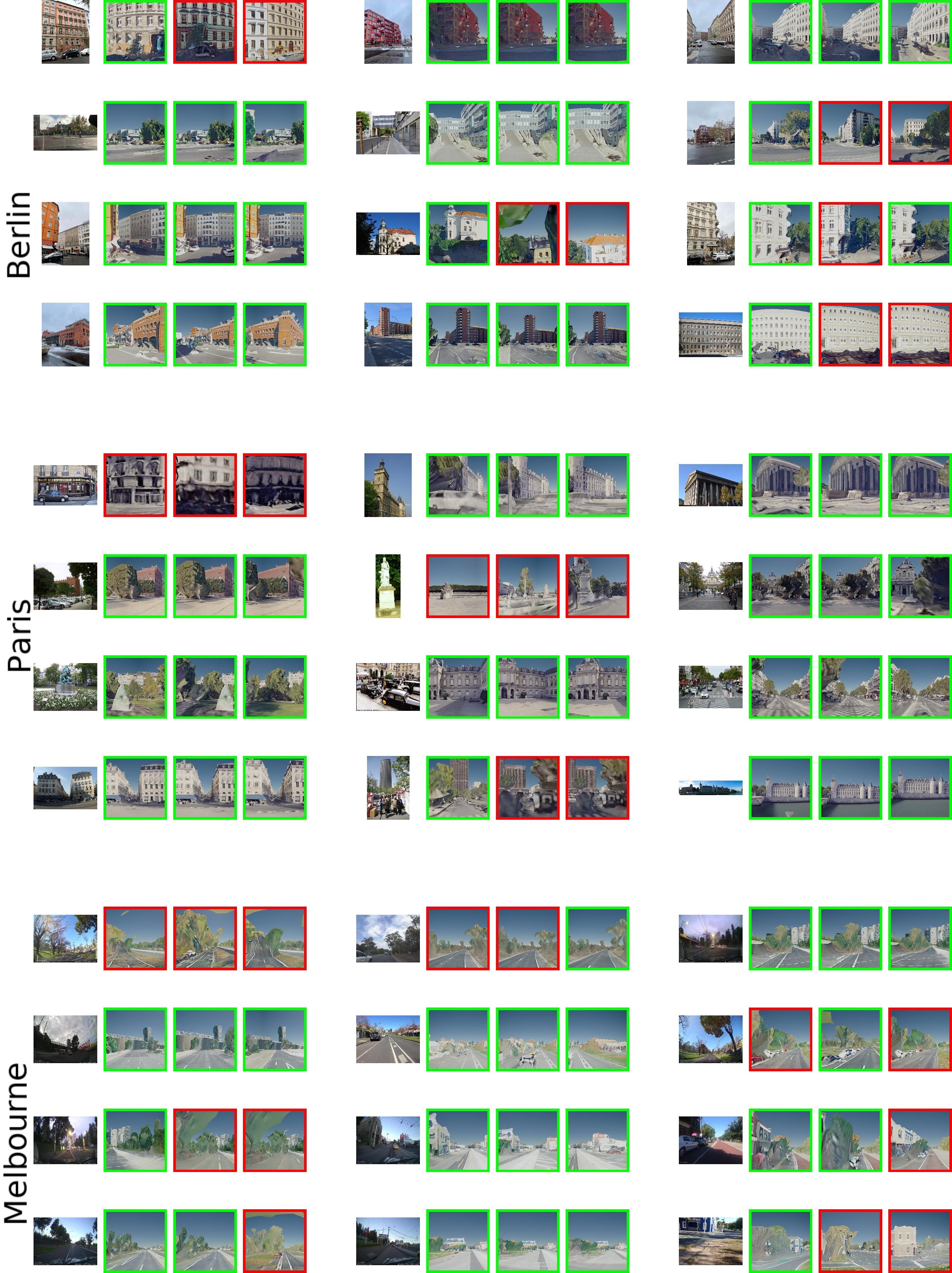}
\caption{\textbf{Qualitative results from the three test sets}, randomly picked, computed with the best model (SALAD + MeshVPR).}
\label{fig:supp_preds_ber_par_mel}
\end{figure*}

Figure \ref{fig:supp_preds_ber_par_mel} shows examples of randomly chosen predictions on synthetic datasets for queries from Berlin, Paris and Melbourne, computed with the best model (SALAD + MeshVPR).
In many cases it is visible how the two models (synthetic and real),
given two images that are semantically similar but visually different, learn to map them nearby in the features space.
For example the trees from the synthetic and real domains look very different, but when a query contains a tree, the predictions often contain a "synthetic" tree.
A similar pattern is visible for statues (see examples from Paris), which are hardly recognizable in the synthetic domain, but are mapped in the same features space by the two models.

